\documentclass[10pt,twocolumn,letterpaper]{article}

\usepackage{iccv}
\usepackage{times}
\usepackage[accsupp]{axessibility}  %

\usepackage{color,xcolor}
\usepackage{epsfig}
\usepackage{graphicx}

\usepackage{cuted}
\usepackage{adjustbox}
\usepackage{array}
\usepackage{booktabs}
\usepackage{colortbl}
\usepackage{float,wrapfig}
\usepackage{hhline}
\usepackage{multirow}
\usepackage{subcaption} %

\usepackage{amsmath,amsfonts}
\usepackage{bm}
\usepackage{nicefrac}
\usepackage{microtype}
\usepackage{inconsolata}
\usepackage{pifont}

\usepackage{changepage}
\usepackage{extramarks}
\usepackage{fancyhdr}
\usepackage{lastpage}
\usepackage{setspace}
\usepackage{soul}
\usepackage{xspace}
\usepackage{indentfirst}
\usepackage{paralist}
\usepackage{dcolumn,booktabs}
\newcolumntype{d}[1]{D{.}{.}{#1}}

\usepackage[pagebackref=true,breaklinks=true,letterpaper=true,colorlinks,citecolor=green,bookmarks=false]{hyperref}
\usepackage{url}

\usepackage{algorithm, algorithmic}
\usepackage{enumerate}
\usepackage{lipsum}

\usepackage{letltxmacro}

\definecolor{citecolor}{HTML}{0071bc}
\definecolor{mydarkblue}{rgb}{0,0.08,1}
\definecolor{mydarkgreen}{rgb}{0.02,0.6,0.02}
\definecolor{mydarkred}{rgb}{0.8,0.02,0.02}
\definecolor{mydarkorange}{rgb}{0.40,0.2,0.02}
\definecolor{mypurple}{RGB}{111,0,255}
\definecolor{myred}{rgb}{1.0,0.0,0.0}
\definecolor{mygold}{rgb}{0.75,0.6,0.12}
\definecolor{mydarkgray}{rgb}{0.66, 0.66, 0.66}

\definecolor{darkblue}{rgb}{0,0.08,1}
\definecolor{darkgreen}{rgb}{0.02,0.6,0.02}
\definecolor{darkred}{rgb}{0.8,0.02,0.02}
\definecolor{darkorange}{rgb}{0.40,0.2,0.02}
\definecolor{darkpurple}{RGB}{111,0,255}

\newcommand{\todocite}[1]{{\color{blue}{[citation needed]}}}

\newcommand{\myparagraph}[1]{\vspace{0pt}\paragraph{#1}}

\newcolumntype{H}{>{\setbox0=\hbox\bgroup}c<{\egroup}@{}}

\definecolor{purple}{RGB}{160, 32, 240}
\definecolor{washblue}{RGB}{186, 224, 228}
\definecolor{sky}{RGB}{128, 128, 128}
\definecolor{seagreen}{RGB}{60, 179, 113}
\definecolor{building}{RGB}{128, 0, 0}
\definecolor{road}{RGB}{128, 64, 128}
\definecolor{sidewalk}{RGB}{0, 0, 192}
\definecolor{fence}{RGB}{64, 64, 128}
\definecolor{vegetation}{RGB}{128, 128, 0}
\definecolor{car}{RGB}{64, 0, 128}
\definecolor{sign}{RGB}{192, 128, 128}
\definecolor{pedestrian}{RGB}{64, 64, 0}
\definecolor{cyclist}{RGB}{0, 128, 192}

\def\be {\begin{equation}}
\def\ee {\end{equation}}
\def\beas {\begin{eqnarray*}}
\def\eeas {\end{eqnarray*}}
\def\bea {\begin{eqnarray}}
\def\eea {\end{eqnarray}}
\def\bes {\begin{equation*}}
\def\ees {\end{equation*}}

\newcommand{\bx}{\mathbf{x}}
\newcommand{\by}{\mathbf{y}}

\newcommand{\bz}{\mathbf{z}}

\newcommand{\bc}{\mathbf{c}}

\newcommand{\cX}{{\cal X}}
\newcommand{\cY}{{\cal Y}}

\usepackage{amssymb}%
\usepackage{pifont}%

\makeatletter
\usepackage{xspace}
\def\@onedot{\ifx\@let@token.\else.\null\fi\xspace}
\DeclareRobustCommand\onedot{\futurelet\@let@token\@onedot}

\def\eg{\emph{e.g}\onedot} 
\def\ie{\emph{i.e}\onedot} 
 
\def\etc{\emph{etc}\onedot}

\makeatother

\iccvfinalcopy 
\def\iccvPaperID{10171}

\ificcvfinal\fi

\begin{document}

\title{MapPrior: Bird's-Eye View Map Layout Estimation with Generative Models}

\author{
Xiyue Zhu$^1$ \hspace{5mm} Vlas Zyrianov$^1$ \hspace{5mm} Zhijian Liu$^2$ \hspace{5mm} Shenlong Wang$^1$ \\
\textsuperscript{1}University of Illinois at Urbana-Champaign \hspace{5mm} \textsuperscript{2}MIT \\\\
\url{https://mapprior.github.io}
}

\maketitle

\ificcvfinal\thispagestyle{empty}\fi

\begin{abstract}

Despite tremendous advancements in bird's-eye view (BEV) perception, existing models fall short in generating realistic and coherent semantic map layouts, and they fail to account for uncertainties arising from partial sensor information (such as occlusion or limited coverage). In this work, we introduce \textbf{MapPrior}, a novel BEV perception framework that combines a traditional discriminative BEV perception model with a learned generative model for semantic map layouts. Our MapPrior delivers predictions with better \textbf{accuracy}, \textbf{realism} and \textbf{uncertainty awareness}. 
{We evaluate our model on the large-scale nuScenes benchmark. At the time of submission, MapPrior outperforms the strongest competing method, with significantly improved MMD and ECE scores in camera- and LiDAR-based BEV perception. }
Furthermore, our method can be used to perpetually generate layouts with unconditional sampling.

\end{abstract}

\section{Introduction}

Accurately understanding the surrounding environment of autonomous vehicles is crucial to guarantee the safety of riders and other traffic participants. 
Among various perception approaches,  Bird's-Eye View (BEV) perception has drawn significant attention in recent years thanks to its capacity to densely model scene layouts and its tight coupling with downstream planning~\cite{centerpoint,LSS,bev_fusion}.

\begin{figure}
\includegraphics[width=\linewidth]{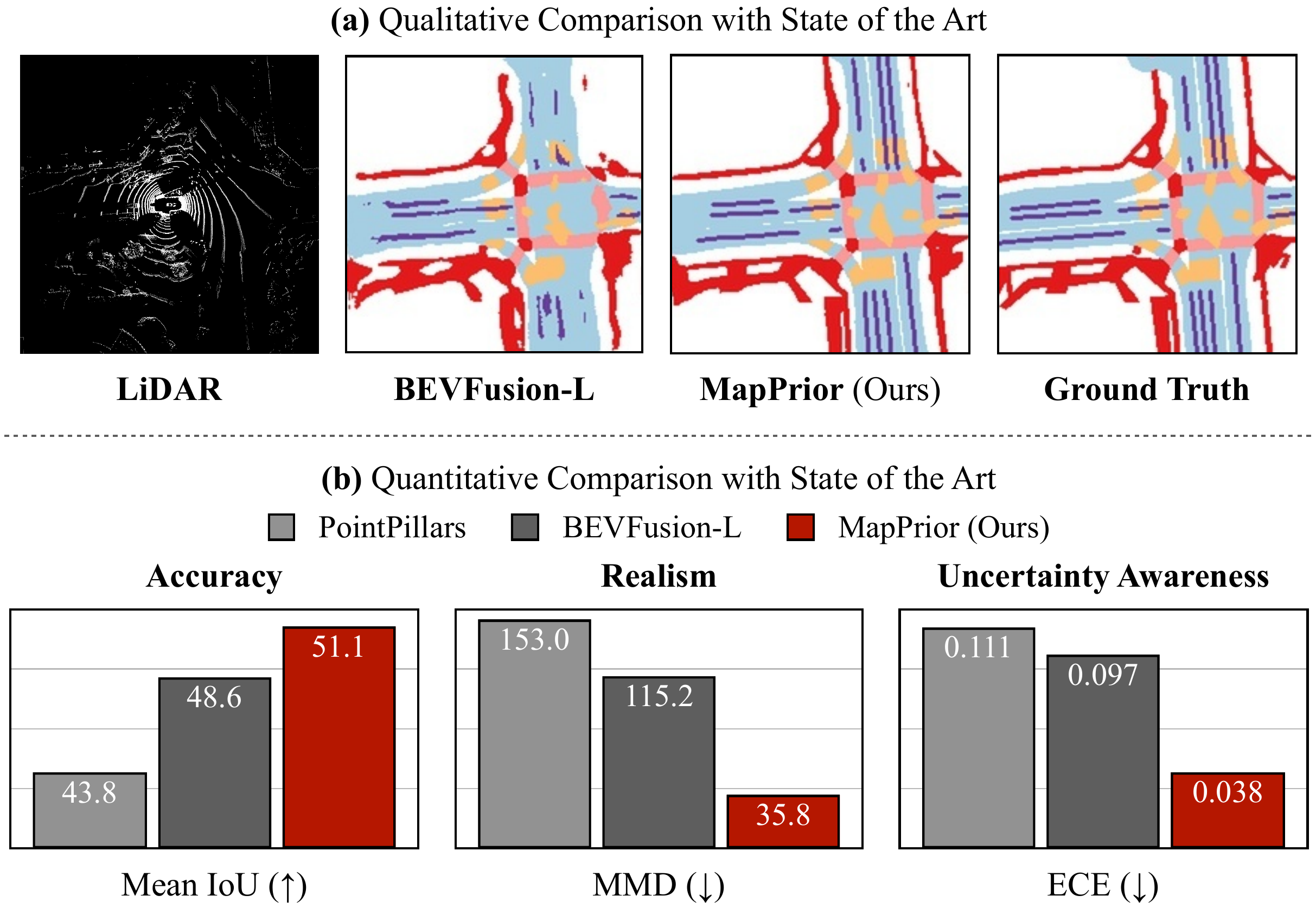} 
\captionof{figure}{Existing predictive BEV perception models do not provide realistic scene structures (\eg, the lane and sidewalks have gaps and are not straight). In contrast, our MapPrior is able to accurately recover the layout with a learned prior. These results are validated with a quantitative comparison, showing our method's superior accuracy, realism and uncertainty awareness. } 
\label{choppy}
\end{figure}

Existing perception models encounter two challenges. The first challenge pertains to the limitations of observations, particularly in distant or occluded regions, resulting in inaccurate predictions. This may manifest in choppy, out-of-distribution, or missing map elements. The second challenge is that most existing models do not consider uncertainty and diversity in possible road layouts.  Taking Fig.~\ref{choppy} as an example, the state-of-the-art LiDAR perception model~\cite{centerpoint} generates incoherent lane markings and sidewalks with massive gaps and cannot quantify the uncertainty and multi-modality as it only produces a single prediction per input.

This paper presents MapPrior, a novel BEV perception method that is accurate, realistic, and uncertainty-aware. At the heart of our method is a novel combination of the standard discriminative BEV perception model with a learned deep generative traffic layout prior. Incorporating generative modeling in this predictive task attempts to address the two aforementioned challenges -- modeling the data distribution improves realism, and using a sampling process allows generating multiple realistic predictions. Combining our generative model with a discriminative perception model ensures that our method retains a strong predictive ability.

Our approach comprises two steps, namely the prediction and generative steps. In the prediction step, we use an off-the-shelf BEV perception model \cite{bev_fusion} to make an initial layout estimate of the traffic scene from sensory input. 
In the generative step, we use our MapPrior model and initial estimate to sample one or multiple refined layouts.  We perform sampling in a learned discrete latent space using a conditional transformer that is provided with the initial prediction. Finally, the generated tokens are passed into a decoder to output the final layout prediction, which is diverse, realistic, and coherent with the input. The encoder, decoder, and codebook of the MapPrior are trained from real-world map data {in an unsupervised way}.

We benchmark our method on the nuScenes dataset against various state-of-the-art BEV perception methods with varying modalities. Our results show that MapPrior outperforms existing methods in terms of accuracy (as reflected by mean intersection-over-union), realism (as reflected by maximum-mean discrepancy), and uncertainty awareness (as reflected by expected calibration error). Furthermore, we demonstrate the unconditional generation capabilities of MapPrior by generating a realistic and consistent HD map of a 30 km-long road.

\section{Related Work}

\begin{figure*}[t] \centering 
\includegraphics[width=\textwidth]{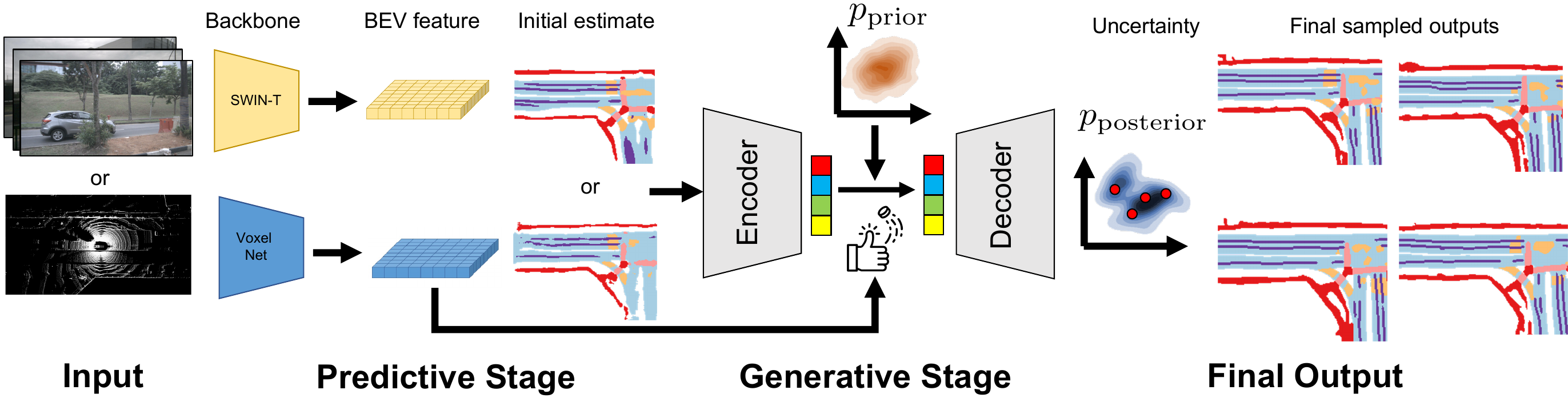} 
\caption{MapPrior first makes use of an off-the-shelf perception model to generate an initial noisy estimate from the sensory input. It then encodes the noisy estimate into a discrete latent code using a generative encoder and generates various samples through a transformer-based controlled synthesis. Finally, MapPrior decodes these samples into outputs with a decoder.}
\label{fig:overview}
\end{figure*}

\noindent\textbf{Self-driving perception}~aims to interpret sensory input and establish a representation of the surrounding environment from onboard sensors. It is a critical component for ensuring the safety and efficiency of autonomous vehicles, and various methods have been proposed to improve accuracy and robustness~\cite{Stanley, juniorcar}. Traditional approaches for self-driving perception involve multiple (isolated) subtasks, such as localizing against a pre-scanned environment \cite{juniorlocalization, barsan18}, object detection and tracking~\cite{poinpillar, centerpoint, pointrcnn, contfuse, fastandfurious}, image segmentation~\cite{unet, pspnet, maskformer} and trajectory prediction~\cite{socialgan, trajectronpp, tnt, lanegcn, intentnet, dsdnet}.

Recently, there has been a growing interest in a unified Bird's Eye View (BEV) perception, both in academia \cite{OFT, CVT, LSS, M2BEV, mp3, fastandfurious} and industry \cite{fiery}. This approach aims to produce a top-down semantic layout of the traffic scene from sensory input, which is efficient, informative, and compact. Notably, the top-down layout is closely linked to downstream 2D motion planning for wheeled robots, making BEV perception particularly suitable for self-driving navigation.

Various BEV perception modules have been studied in robotics and computer science to improve perception accuracy. These approaches typically take 
LiDAR~\cite{centerpoint, poinpillar, mao2021voxel, shi2023pv}, multi-view 
camera~\cite{OFT, CVT, LSS, M2BEV, chitta2021neat, liu2022petr, fiery, reading2021categorical, gosala22bev}, or 
both~\cite{MVP, PointPainting, bev_fusion, transfusion, Prakash2021CVPR} 
as input, and output segmentations for road elements and detections for dynamic traffic participants like cars, cyclists or pedestrians. The majority of these methods are supervised and rely on predictive networks like 
CNNs \cite{reading2021categorical, shi2023pv, centerpoint, poinpillar, contfuse, OFT, LSS} and 
transformers \cite{transfusion, Prakash2021CVPR, chitta2021neat, liu2022petr, mao2021voxel, CVT, gosala22bev, fiery, huang2021bevdet}. Despite their success, there are still unresolved challenges. For example, real-world traffic layouts are highly structured, with straight lane marks, strong topological relationships, and sharp road element boundaries. But such structures have proven difficult to preserve in the BEV perception output map, even with state-of-the-art methods, as shown in Fig. \ref{choppy}. This can harm the realism of perception output and the practicality of the resulting map for motion planning. Additionally, most networks make a single layout estimation without diversity or calibrated uncertainty, leaving the autonomy vulnerable to catastrophic failure.

HDMapNet~\cite{hdmapnet}, VectorMapNet \cite{vectormapnet}, and others \cite{Mattyus_2016_CVPR, mp3} have been proposed to address the issue of layout structure by introducing a vectorized output map format with implicitly structured priors. However, this approach fails to account for all layer types and still produces significant artifacts in regions with limited observations. Very recently, calibrated confidence scores have been investigated to address the aforementioned uncertainty issue \cite{calibratedlocalization, motionplanningreview},
but the inherent multi-modal uncertainty in BEV perception remains under-explored. 

In contrast, our approach leverages generative modeling as a map prior, which encodes the rich structure of the traffic scene. Our conditional generative modeling also allows us to sample multiple diverse outputs for the same input. Consequently, our results are more accurate and realistic with better multi-modal uncertainty modeling.

\myparagraph{Generative models.}

Our approach to generative BEV perception involves learning a map prior and performing conditional sampling using deep generative models. Generative models learn to capture the underlying structure of the data distribution and create new, diverse samples. Several well-known approaches include VAEs (variational autoencoders)~\cite{VAE,NVAE}, which use variational inference to learn a latent space model, and autoregressive models \cite{Image_Transformer}, which decompose the generation problem into simpler conditional generation tasks. GANs (generative adversarial networks)~\cite{GAN,attentionGAN,WGAN,styleGAN,pix2pix,regu_gan,bigGAN} use an adversarial loss to train networks to convert noise into samples. Flow-based models~\cite{GLOW,nice_flow} exploit an invertible process to sample from a proposal distribution. Recently, diffusion models~\cite{diffusion,DDPM} have been developed that generate samples through a denoising-diffusion process. Our method leverages vector-quantized generative models~\cite{VQVAE, VQGAN, maskgit2}, which use a discrete-valued latent space representation that is powerful, structure-preserving, and efficient for conditional sampling. Specifically, we build our map prior on top of this approach to capture the discrete structure of the map data and generate high-quality samples.

\myparagraph{Generative approach for vision.}
Our approach belongs to the broader category of generative modeling approaches in computer vision~\cite{inpaintingdeepgm, discvsgen, Ranzato11, srivastava12}, which includes Markov random fields (MRFs), factor graphs, energy-based models, deep generative models, among others. These practices date back to the 1970s. In contrast to discriminative or predictive approaches, generative approaches often tackle the task as a marginal sampling or MAP inference problem, which is more effective at utilizing strong prior knowledge and modeling uncertainty. Representative works can be seen in low-level vision~\cite{lllv, frame}, optical flow~\cite{secretofflow, fieldofexpert}, and image segmentation~\cite{normalizedcut, densecrf, crfasrnn}.
Generative approaches with conditional sampling have often been investigated in vision tasks involving multimodality or uncertainty, such as image editing~\cite{cyclegan, pix2pix, jojogan, swappingae}, image segmentation~\cite{semanticGAN} and stereo estimation~\cite{deeppruner, contmrf}. However, few studies have explored their use in self-driving perception, despite its inherent multi-modal uncertainty nature.

\section{Approach}
\label{sec:method}

The objective of this research is to develop a method that can generate a \textit{precise}, \textit{realistic}, and \textit{uncertainty-aware} map layout from sensory input in one or a few modalities. To achieve this, we introduce a new framework named MapPrior that combines the predictive capability of discriminative models with the capacity of generative models to capture structures and uncertainties. We present a two-stage conditional sampling framework to explain our approach and detail the implementation of each module. Additionally, we discuss our design decisions, the learning process, and how our approach relates to previous methods.

\subsection{Overview}
\label{sec:overview}

\paragraph{Formulation.}

We formulate the probabilistic BEV perception problem as a conditional sampling task. Given the sensory input $\bx \in \cX$ (which could be from a camera, LiDAR or multiple sensors), we aim to find one or multiple plausible traffic layouts $\by \in \cY $ from the top-down bird's-eye view. The traffic layout $\by$ is a multi-layer binary map centered at the ego vehicle.
Conventional methods~\cite{bev_fusion, centerpoint, LSS, MVP, M2BEV} rely on a deterministic predictive network $\by = f_\theta(\bx)$ to provide a single output estimation. We propose, instead, to use a generative vector-quantized latent space model to model the uncertainty and diversity. Specifically, we aim to design a conditional probability model $p_\theta(\by | \bx)$ that can sample our desired output $\by$ from a distribution. This allows us to explicitly model uncertainty and generate multiple plausible traffic layouts. 

\myparagraph{Motivation.}

Intuitively speaking, the objective of a predictive perception model is to %
produce a coherent semantic layout that best represents the sensory input. However, this training objective  
does not necessarily optimize for structure preservation or realism. For example, minor defects in lane markers might not significantly affect cross-entropy, but could drastically impact downstream modules due to topological changes. On the other hand, a generative prior model, such as a GAN, is trained to capture realism in structures. The advantage of such a model is that it can be trained on HD map collections, without paired data, in an unsupervised manner. This insight inspires our proposed solution, which combines a predictive model with a generative model to tackle the conditional sampling task with both coherence and realism in mind.

\subsection{Inference}
\label{sec:inference}

In our framework, the distribution $p_\theta(\by | \bx)$ is defined implicitly using a latent model $\bz$. 
The framework comprises of a predictive stage and a generative stage. During the predictive stage, the perception module $F(\bx)$ generates an initial noisy estimate $\by^\prime$ for input $\bx$. 
For the generative stage, we take inspiration from recent successful techniques in vector-quantized generation~\cite{VQGAN, maskgit2, stablediffusion} for conditional sampling and use a VQGAN-like model to generate multiple realistic samples. To achieve this, we encode the noisy estimate $\by^\prime$ into a discrete latent code $\bz^\prime$ using a generative encoder $E(\by^\prime)$. This latent code $\bz^\prime$ and the sensory input $\bx$ then guide the generative synthesis process through a transformer-based controlled synthesis in the latent space, producing various samples $\{\bz^{(k)} \sim p(\bz | \bz^\prime, \bx)\}$. Finally, these samples are decoded into multiple output samples using a decoder $G(\cdot)$: ${\by^{(k)} = G(\bz^{(k)})}$, which provides our final layout estimation samples. Fig.~\ref{fig:overview} depicts the overall inference framework.

\myparagraph{Predictive stage.}

The predictive stage aims to establish reliable initial layout estimation that can act as a guiding control during the conditional sampling stage. To achieve this, we have incorporated a predictive sensory backbone, {VoxelNet~\cite{SECOND,centerpoint}} for LiDAR and Swin Transformer~\cite{liu2021swin} for multi-view camera inputs. The sensory backbone networks first extract features from the sensory space and then project them into bird's eye view features. For the LiDAR input, a 3D convolution backbone is utilized as a feature extractor, and the LiDAR features are flattened along the height dimension to project them to BEV. In contrast, for the camera input, a hierarchical transformer is used as a feature extractor for each view, and monocular depth estimation and view transformation are applied to project the features to BEV. The predictive model employs a convolutional segmentation head to generate the layout estimates, denoted by $\mathbf{y}^\prime = F(\mathbf{x})$. It should be noted that the resulting BEV map achieves a reasonable intersection-over-union score (IoU). However, as Fig.~\ref{choppy} highlights, the model suffers from unrealistic structures, missing road elements, and noise, particularly in regions with limited observations.

\myparagraph{Generative stage.}

To enhance the quality and diversity of our perception, we incorporate a generative map prior in the second stage for conditional generation. The generative map prior is built on a VQGAN architecture~\cite{VQGAN}, which consists of three learnable components: the encoder $E$, the decoder $G$, and the codebook $\mathcal{C} = \{ \bc_j \}$ with $j$ being the code index. The encoder transforms a traffic layout into a latent feature map, where each element of the spatial map is chosen from one of the codes in the codebook as follows:
\begin{equation}
\label{eq:encode}
\bz_t(\by) = \mathop{\arg\max}_{\bc_j \in \mathcal{Z} }  \| \bc_j - E_t(\by)\|_2^2, \quad \forall t
\end{equation}
where $t$ is the $t$-th entry of the feature map $\bz$. The decoder then transforms the latent code back to the layout map space: $G(\mathbf{z}(\by))$. Sampling from this prior can be done by randomly drawing a latent code and decoding it into a layout map. The discrete-valued auto-encoder architecture greatly regularizes the output space in a structured manner, preventing it from producing unrealistic reconstructions, as shown in Fig~\ref{choppy}.

During conditional sampling at inference, we first encode the noisy estimate $\by^\prime$ into a discrete latent code $\bz^\prime$ using a generative encoder $E$. This latent code $\bz^\prime$ and the sensory input $\bx$ are then used as guidance by a transformer $T(\bz^\prime, \bx)$ for the generative synthesis of the latent space, producing various samples ${\bz^{(k)} \sim p(\bz | \bz^\prime, \bx) = T(\bz^\prime, \bx)}$. 
Specifically, we use an autoregressive scheme to progressively sample each code, \ie $p(\bz | \bz^\prime, \bx) = \prod_t p(\bz_t | \bz_{<t}^{(k)}, \bz^\prime, \bx)$. At the $t$-th step, the transformer takes as input the current latent code $\bz_{<t}$, the guidance code $\bz^\prime$ as well as encoded sensory feature $\bx$ as input tokens, and outputs the next token's probability over the codebook $p(\bz_t | \bz_{<t}^{(k)}, \bz^\prime, \bx)$:
\begin{equation}
\label{eq:autoregressive}
{\bz_t^{(k)} \sim p(\bz_t | \bz_{<t}^{(k)}, \bz^\prime, \bx)}
\end{equation}
where $p(\bz_t | \bz_{<t}^{(k)}, \bz^\prime, \bx)$ is the conditional probability estimated from the transformer.  We use nucleus sampling \cite{NucleusSampling} to get multiple diverse $\bz^{(k)}$, which trades-off between sampling quality and diversity. Finally, these samples are decoded into multiple output samples using a decoder $G(\cdot)$: ${\by^{(k)} = G(\bz^{(k)})}$, which provides our final layout estimation samples. Formally speaking, the entire second stage can be written as follows:
\begin{equation}
\label{eq:sampling}
\by^{(k)} = G(\bz^{(k)}) \text{\ \ where\ \ } \bz^{(k)} \sim p(\bz | \bz^\prime, \bx).
\end{equation}

\myparagraph{One-step generation.}

{To enhance inference speed, we introduce a one-step variant of MapPrior, which generates a single sample rather than multiple diverse ones. It also produces tokens in a single step, bypassing the autoregressive sampling strategy:
\begin{equation}
\label{eq:sampling-one-step}
\by = G(\bz) \text{\ \ where\ \ } \bz \sim p(\bz | \bx).
\end{equation}
This provides an effective way to trade off between the generation quality and efficiency.}

\subsection{Learning} 

The training process consists of three individual components. Firstly, the perception module, denoted as $F$, is learned to produce a reliable initial layout estimation. Secondly, the encoder, decoder, and codebook (denoted as $E$, $G$, and $\mathcal{C}$, respectively) are jointly trained to represent a strong map prior model. Lastly, given a fixed map prior model, the conditional sampling transformer $T$ is learned to sample high-quality final results. In the following, we will provide a detailed description of each component's training procedure.

\myparagraph{Training the perception module.}

To train the perception module $F$, we follow the standard practice adopted in prior works~\cite{bev_fusion, LSS} and employ a binary cross-entropy loss. The objective function is defined as follows: $\min_F - \left[ \by_{\mathrm{gt}} \log \by^\prime \right]$, where $\by_{\mathrm{gt}}$ denotes the ground-truth label, and $\by^\prime = F(\bx)$ is the corresponding prediction given the input $\bx$.

\begin{figure*}[h] \centering
\includegraphics[width=\textwidth]{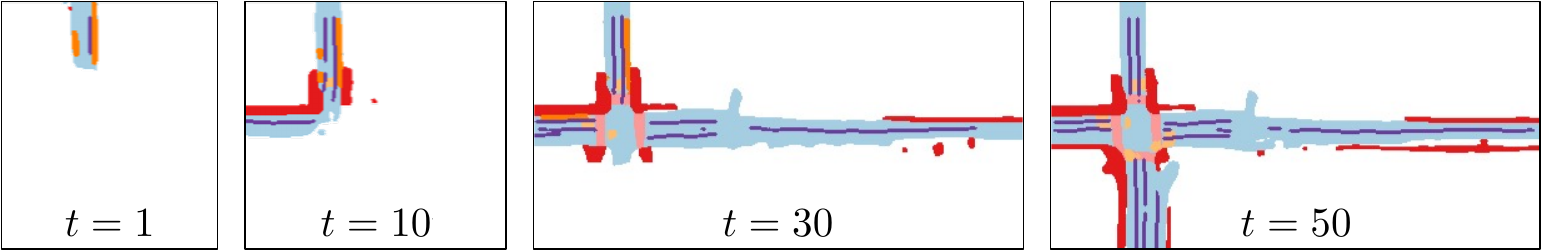}
\caption{MapPrior can be exploited in a progressive manner to generate perpetual traffic layouts. At each step, we choose a local area to sample to fill unexplored regions and expand the frontiers.}
\label{fig:perpetual}
\end{figure*}

\myparagraph{Training the map prior model.}

The training procedure for the map prior model involves jointly optimizing the encoder $E$, decoder $G$, and the fixed-size codebook $\mathcal{C}$ in an end-to-end fashion. We use a vector-quantized auto-encoder to reconstruct the input data $\by$ through $\hat{\by} = G(\bz(\by_i))$ following Eq.~\ref{eq:encode}. The following losses are minimized:
\begin{equation} \label{eq:vq} \min_{E, G, \mathcal{C}} \max_D \mathcal{L}_\mathrm{recon} + \mathcal{\lambda}_\mathrm{GAN} \mathcal{L}_\mathrm{GAN} + \mathcal{L}_\mathrm{latent}. \end{equation}

The first term, $\mathcal{L}_\mathrm{recon}$, is the reconstruction loss, which is designed to maximize the agreement between the input $\mathbf{y}$ and the reconstructed output:
\begin{equation}
    \mathcal{L}_\mathrm{recon} = \| \hat{\mathbf{y}} - \mathbf{y} \|^2.
\end{equation}

In addition to the reconstruction loss, the second term is a GAN loss inspired by VQGAN~\cite{VQGAN} to encourage the reconstructed output to be realistic with a clear local structure and topology. A local discriminator $D(\cdot)$ is trained to differentiate between real and reconstructed BEV layout using cross-entropy loss, and the $\mathcal{L}_\mathrm{GAN}$ loss aims to make the reconstructed image as realistic as possible through fooling the discriminator:
\begin{equation}
\mathcal{L}_\mathrm{GAN} = \log D(\by) + \log (1 - D(\hat{\by})),
\end{equation}
where $\by$ is a real sample and $\hat{\by}$ is a reconstructed sample. {Following VQGAN~\cite{VQGAN}, this loss is rescaled by
\begin{equation}
\mathcal{\lambda}_\text{GAN} = g_\text{rec}/(g_\text{GAN}+\sigma),    
\end{equation}
where $g_\text{rec}$ and $g_\text{GAN}$ are gradients of $\mathcal{L}_\mathrm{rec}$ and $\mathcal{L}_\mathrm{GAN}$ with respect to the last layer of the generator.}

The last term promotes the expressiveness of the codebook. We expect the latent feature $E(\by)$ to be close to its nearest codebook token and vice versa. Thus, the latent loss is defined as
\begin{equation}
\mathcal{L}_\mathrm{latent} = \| \mathbf{z}- \mathrm{sg}[E(\mathbf{y})] \|^2 + \| \mathrm{sg}[\mathbf{z}]- E(\mathbf{y}) \|^2,
\end{equation}
where $\mathrm{sg}[\cdot]$ is a gradient detach operator.

\myparagraph{Training the conditional sampler.}

Our conditional sampling transformer, denoted as $T$, is trained in the latent code space to sample high-likelihood latent codes given input controlling guidance. Given a paired input and output $(\bx, \by)$ and fixed perception module $F$, map prior model $E, G, \mathcal{C}$, the transformer is trained using the following objective:
\begin{equation} \min_{T} \mathcal{L}_\mathrm{CE} + \mathcal{L}_\mathrm{out}.\end{equation}

The first loss, $\mathcal{L}_\mathrm{CE}$, optimizes the transformer to maximize the estimated probability of the latent code of the ground-truth layout. We use a cross-entropy loss between the transformer's estimated probability $T(\bz^\prime, \bx)$ and the ground-truth sample $\by$:
\begin{equation} 
\mathcal{L}_\mathrm{CE} = \sum_t \sum_i \by_{i, t} \log T_{i, t}( \bz^\prime,\bx),
\end{equation}
where, $i$ represents the $i$-th output label, and $t$ represents the $t$-th autoregressive step. 

Although making the latent code closer to the ground-truth map's latent code is essential, it is not sufficient to ensure high output fidelity. Hence, we include an additional reconstruction output loss that encourages the transformer to favor samples that produce high-accuracy layouts. Similar to the reconstruction loss in auto-encoding training, an L2 reconstruction loss is used:
\begin{equation} 
\mathcal{L}_\mathrm{out} = \| \mathbf{y} - \mathbf{\hat{y}} \|_2^2,
\end{equation}
where $\mathbf{y}$ is the true map and $\mathbf{\hat{y}}$ is the predicted map. Note that codebook selection in latent code space is non-differentiable; thus, we use Gumbel-Softmax \cite{jang2016categorical} to ensure differentiability in practice throughout the training processes.

\begin{table*}[!t]
\caption{Quantitative results of BEV map segmentation on nuScenes. Our MapPrior achieves better accuracy (IoU), realism (MMD) and uncertainty awareness (ECE) than discriminative BEV perception baselines.}
\setlength{\tabcolsep}{5pt}
\small\centering
\begin{tabular}{lcccccccccc}
\toprule    
 & \multirow{2.5}{*}{Modality} & \multicolumn{7}{c}{IoU ($\uparrow$)} & \multirow{2.5}{*}{MMD ($\downarrow$)} & \multirow{2.5}{*}{ECE ($\downarrow$)} \\
 \cmidrule(lr){3-9}
 & & Drivable & Ped.~X & Walkway & Stop Line & Carpark & Divider & Mean \\
\midrule
OFT \cite{OFT}  & C & 74.0 & 35.3 & 45.9 & 27.5 & 35.9 & 33.9 & 42.1 & 54.5 & 0.045 \\ %
LSS \cite{LSS}  & C & 75.4 & 38.8 & 46.3 & 30.3 & 39.1 & 36.5 & 44.4 & 43.2 & 0.041\\ %
BEVFusion-C~\cite{bev_fusion}     & C & \textbf{81.7} & \textbf{54.8} & \textbf{58.4}  & \textbf{47.4}   & 50.7   & \textbf{46.4} & 56.6  & 39.6 & 0.038 \\
\textbf{MapPrior-C}  & C   & \textbf{81.7} & 54.6 & 58.3 & 46.7 & \textbf{53.3} & 45.1 &  \textbf{56.7} & \textbf{28.4} & \textbf{0.026} \\ 
\textcolor{black}{\textbf{MapPrior-C} (1 step)} & C   & {81.6} & 54.6 & \textbf{58.4} & 46.8 & \textbf{53.9} & 45.1 &  {56.7} & {28.7} & -- \\ 
\midrule
PointPillars~\cite{poinpillar}    & L & 72.0 & 43.1 & 53.1 & 29.7 & 27.7 & 37.5 & 43.8 & 153.0 & 0.111 \\
BEVFusion-L~\cite{centerpoint,bev_fusion}     & L & 75.6 & 48.4 & 57.5 & 36.5 & 31.7 & 41.9 & 48.6 & 115.2 &  0.090\\ 
\textbf{MapPrior-L} & L & \textbf{81.0} & \textbf{49.7} & \textbf{58.0} & \textbf{37.5} & \textbf{38.2} & \textbf{42.4} & \textbf{51.1} & \textbf{35.8} & \textbf{0.038} \\
\textcolor{black}{\textbf{MapPrior-L} (1 step)} & L & 80.1 & 49.0 & 57.8 & 37.8 & 33.0 & 42.5 & 50.0 & {50.2} & -- \\
\midrule
{PointPainting~\cite{PointPainting}} & {C+L} & {75.9} & {48.5} & {57.1} & {36.9} & {24.5} & {41.9} & {49.1} & {109.8} & {0.099} \\
{MVP~\cite{MVP}} & {C+L} & {76.1} & {48.7} & {57.0} & {36.9} & {33.0} & {42.2} & {49.0} & {115.3} & {0.096} \\
{BEVFusion~\cite{bev_fusion}} & {C+L} & {85.5} & {60.5} & {67.6} & {52.0} &  {57.0} &{53.7} & {62.7} & \textbf{21.6} &{0.038} \\
\textcolor{black}{\textbf{MapPrior-CL}} & C+L & 85.3 & \textbf{61.4} & 67.1 & \textbf{51.7} &  \textbf{60.0} &53.3 & \textbf{63.1} & 28.0 & \textbf{0.020} \\
\textcolor{black}{\textbf{MapPrior-CL} (1 step)} & C+L & 85.3 & 61.3 & 67.0 & 51.7 &  59.6 &53.1 & 63.0 & 28.1 & -- \\
\bottomrule

\label{tab:IOU}
\end{tabular}
\end{table*}

\subsection{Discussions}

\paragraph{Uncertainty quantification.}

Our latent transformer offers diverse results $\by^{(k)}$ by using a conditional sampling scheme. By estimating the variance of $\by^{(k)}$, we can estimate the uncertainty map for our results. By using the average of $\by^{(k)}$, we aggregate the diverse result samples to a more stable and better calibrated results.

\myparagraph{Perpetual layout generation.}
{Our trained map prior enables the continuous generation of realistic and varied driving sequences, which are extremely valuable for content creation and autonomous driving simulations.} We use a progressive generation strategy, which builds upon prior works~\cite{patchmatch, quilting, infinite_nature_2020, sgam, VQGAN, maskgit2}. The strategy involves iteratively expanding our visual horizon and generating new content by leveraging our map prior model to fill in the previously unseen areas. We illustrate this process in Fig.~\ref{fig:perpetual}.

\section{Experiments}

\begin{figure*}[!t]
\centering
\includegraphics[width=1.035\textwidth]{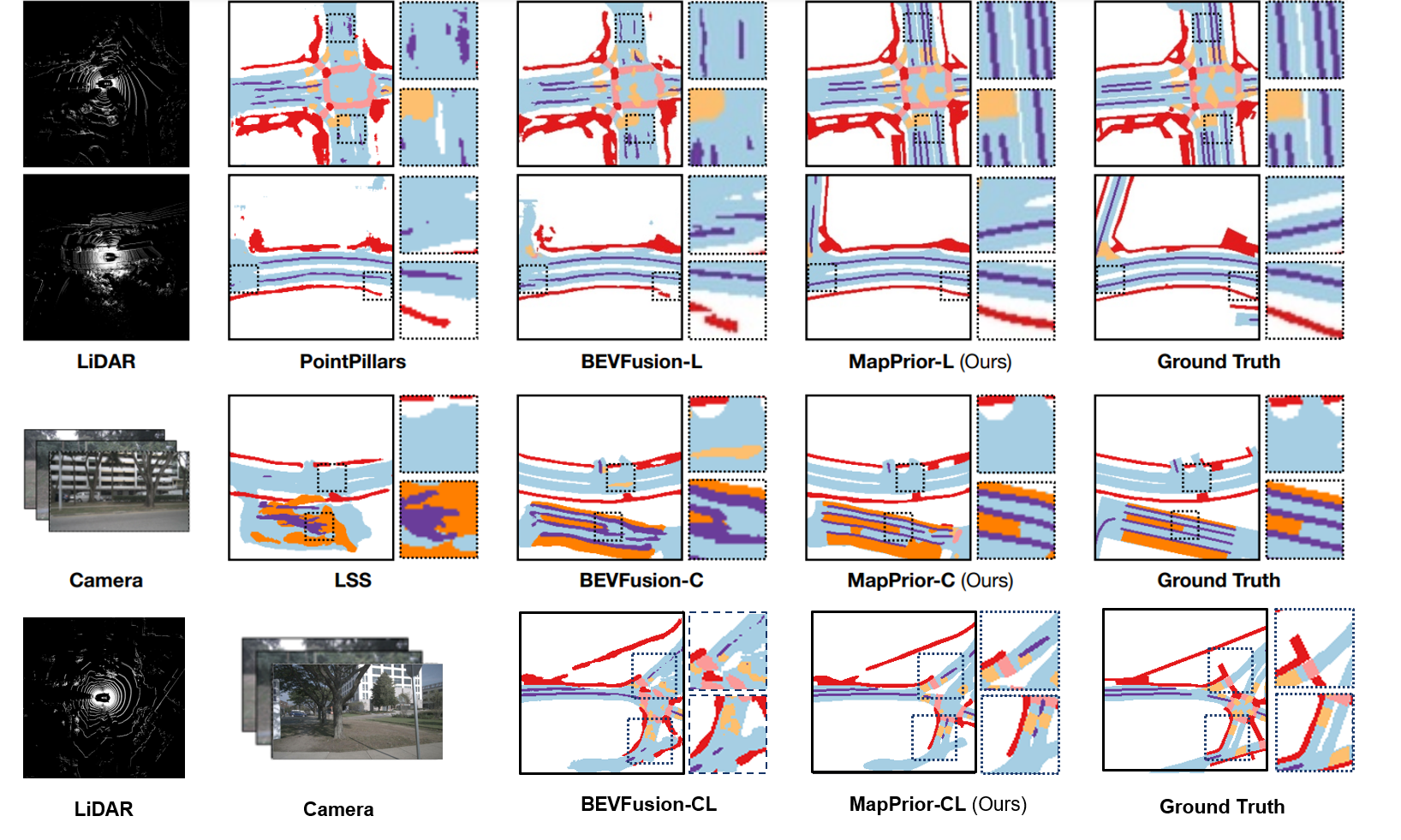}
\captionof{figure}{Qualitative results of BEV map segmentation on nuScenes. Discriminative BEV perception baselines produce results with clear artifacts (\eg, lane markings are discontinuous, roads and pedestrian walkways have unrealistic gaps, \etc). In contrast, our MapPrior produces semantic map layouts that are clean, realistic, and in-distribution.}
\label{qual_results}
\end{figure*}

\begin{figure*}[!t] \centering 
 \begin{tabular}{cccccc} \centering
 \includegraphics[width=0.145\textwidth]{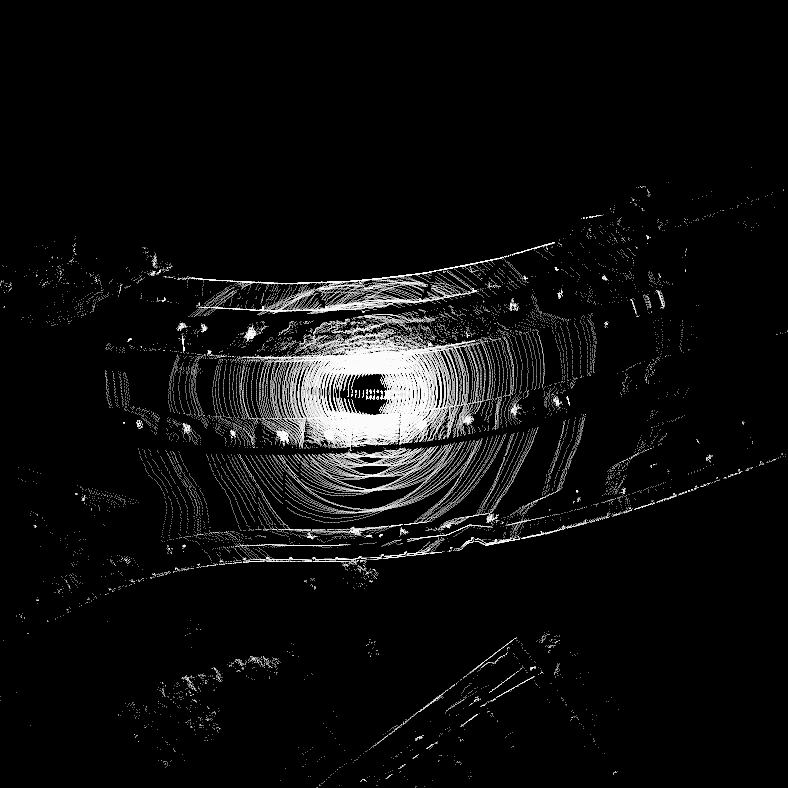}
 &\includegraphics[width=0.145\textwidth]{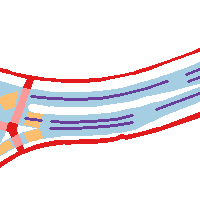}
 &\includegraphics[width=0.145\textwidth]{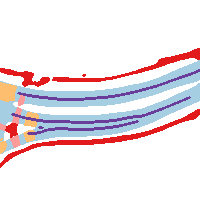}
&\includegraphics[width=0.145\textwidth]{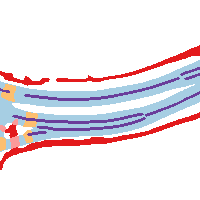}
&\includegraphics[width=0.145\textwidth]{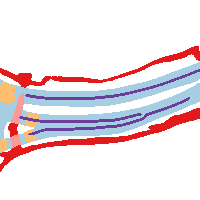}
&\includegraphics[width=0.145\textwidth]{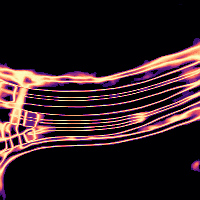} 
\\
\includegraphics[width=0.145\textwidth]{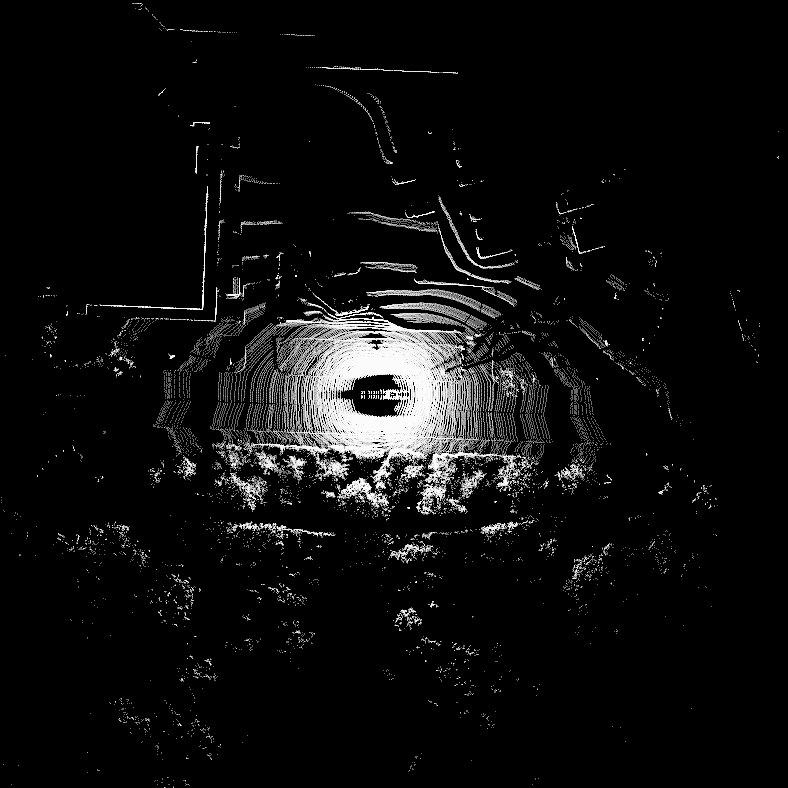}
 &\includegraphics[width=0.145\textwidth]{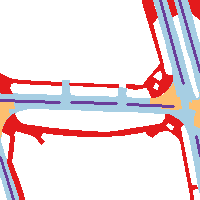}
 &\includegraphics[width=0.145\textwidth]{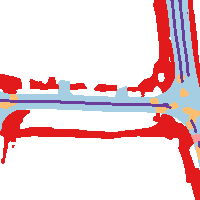}
&\includegraphics[width=0.145\textwidth]{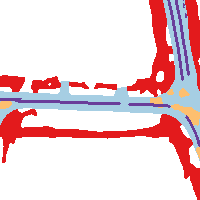}
&\includegraphics[width=0.145\textwidth]{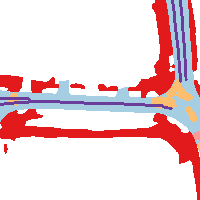}
&\includegraphics[width=0.145\textwidth]{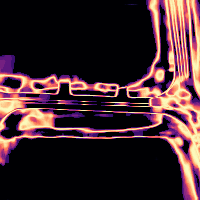}
\\
Input & GT & Our Sample 1 & Our Sample 2 & Our Sample 3 &  Uncertainty Map
 \end{tabular}
 \caption{Qualitative results of diversity on NuScenes.}
 \label{Sampling Diversity}
 \end{figure*}

\begin{figure*}[!t] \centering 
 \begin{tabular}{ccccccc} \centering

\includegraphics[width=0.12\textwidth]{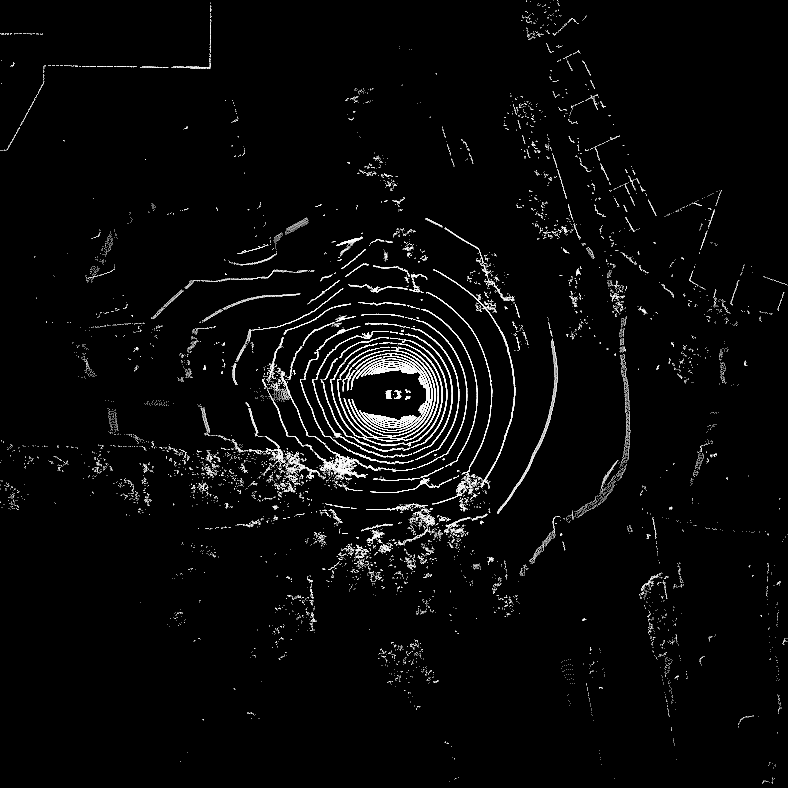} &
 \includegraphics[width=0.12\textwidth]{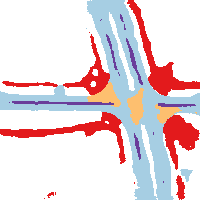} &  
 \includegraphics[width=0.12\textwidth]{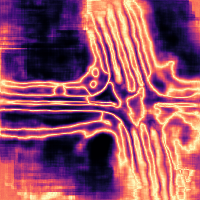} & 
 \includegraphics[width=0.12\textwidth]{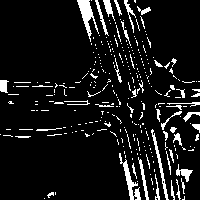} & 
 \includegraphics[width=0.12\textwidth]{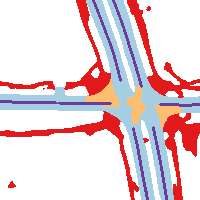} &  
 \includegraphics[width=0.12\textwidth]{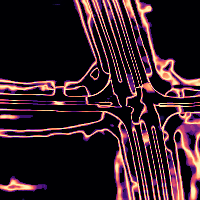} &
  \includegraphics[width=0.12\textwidth]{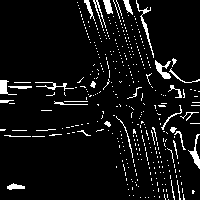} \\

   \includegraphics[width=0.12\textwidth]{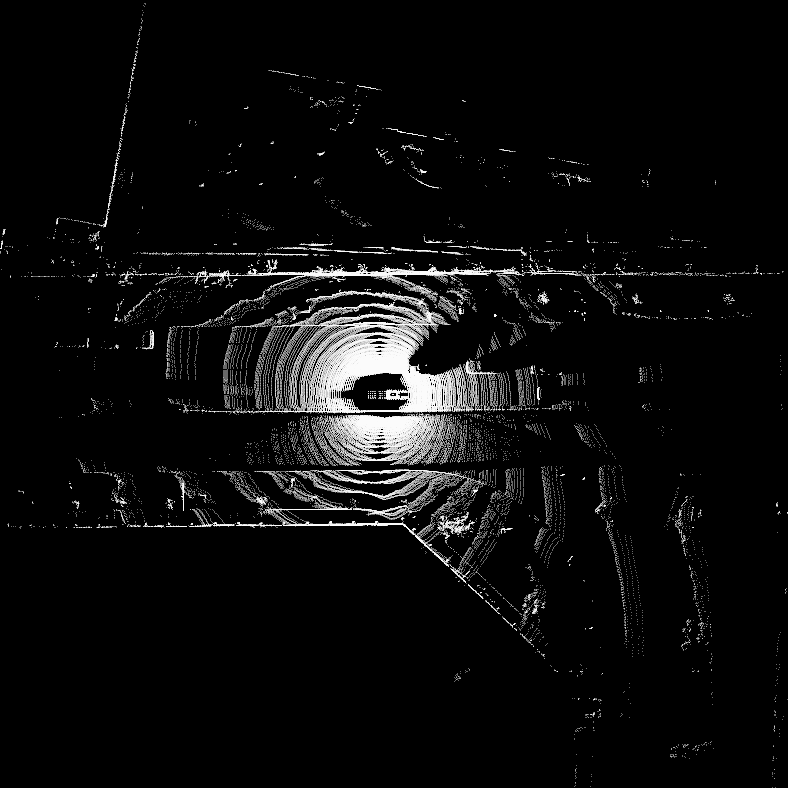} & 
 \includegraphics[width=0.12\textwidth]{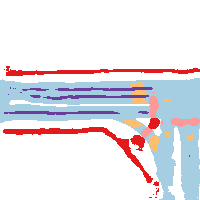} &  
 \includegraphics[width=0.12\textwidth]{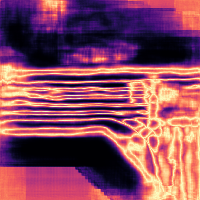} & 
 \includegraphics[width=0.12\textwidth]{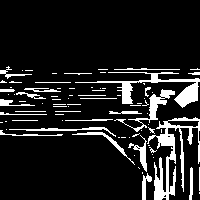} & 
 \includegraphics[width=0.12\textwidth]{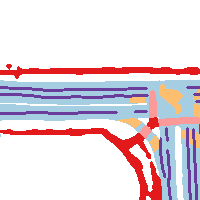} &  
 \includegraphics[width=0.12\textwidth]{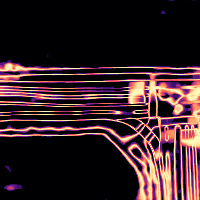} &
  \includegraphics[width=0.12\textwidth]{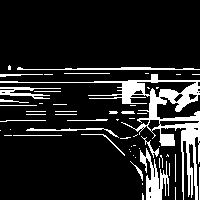} \\

 \multirow{2.5}{*}{LiDAR} &
 Prediction & Uncertainty & Error Map & Prediction & Uncertainty & Error Map 
 
 \\
 \cmidrule(lr){2-4}
 \cmidrule(lr){5-7}
 
 & \multicolumn{3}{c}{BEVFusion-C}
 & \multicolumn{3}{c}{MapPrior}
 \\

 \end{tabular}
 \caption{Prediction, uncertainty, and error map comparison between BEVFusion-C and MapPrior. BEVFusion generates a misguided uncertainty boundary by giving high weight to areas where map elements are unlikely to be. In comparison, MapPrior's uncertainty is constrained to map element boundaries. This is further confirmed with the error map, where MapPrior's uncertainty map closely matches the error map. }
 \label{Uncertainty Calibration}
 \end{figure*}

We evaluate our MapPrior on BEV map segmentation and generation tasks for both LiDAR and camera modalities. We evaluate our approach to generate an accurate and realistic traffic layout both quantitatively and qualitatively in Sec.~\ref{sec:4.2}. We show how output loss and BEV features can affect our performance in Tab.~\ref{tab:abalation}. We finally estimate how our model can generate diverse samples and how our model is calibrated using the diverse samples in Fig.~\ref{Sampling Diversity} and \ref{Uncertainty Calibration}. Specifically, we are interested in seeing how using a generative prior affects accuracy (reflected by mIoU), realism (reflected by MMD), and model calibration (reflected by ECE). 

\subsection{Experimental Setup}

\paragraph{Datasets.} 

We evaluate our model on nuScenes \cite{nuscenes}, a large-scale outdoor autonomous driving dataset containing  1000 driving scenes, consisting of 700 scenes for training, 150 for validation, and 150 scenes for testing. It has around  40,000 annotated key-frames, each with six monocular camera images encompassing a 360-degree FoV (used by camera models), and a 32-beam LiDAR scan (used by LiDAR models). We follow the train/validation split provided by nuScenes. We evaluate all models on the validation set following common practices in BEV layout estimation~\cite{bev_fusion, LSS}. The ground truth segmentation map is provided by nuScenes and was labeled manually \cite{nuscenes}. We rasterized the map layers from the nuScenes map into the ego frame.

\myparagraph{Metrics.} 

We provide quantitative results for both segmentation and generative tasks. For BEV map segmentation, our metric is the Intersection-over-Union (IoU score) for six map classes (drivable area, pedestrian crossing, walkway, stop line, car-parking area, and lane divider), as well as the mean IoU averaged over all six map classes. Since different classes may overlap, we separately compare the IoU score for each class. For all baseline predictive models, we choose the best threshold that maximizes IoU for comparison to ensure no bias is introduced due to the suboptimal threshold selection. 
For MapPrior, we simply use 0.5 as the threshold.

To evaluate generated map layout realism, we use the maximum mean discrepancy (MMD) metric. MMD effectively measures the distance between two distributions of two sets of samples by measuring the squared difference of their mean in different spaces: 
\begin{equation}
\begin{aligned}
    \text{MMD} = \sum_i^n \sum_{i^\prime}^n k(\bx_i, \bx_{i^\prime}) / n^2 + \sum_j^m \sum_{j^\prime}^m k(\bx_j, \bx_{j^\prime}) / m^2 \\
    - 2 \cdot {\sum_i^n \sum_{j}^m k(\bx_i, \bx_{j})} / nm.
\end{aligned}
\end{equation}
To evaluate a set of predicted layouts, we compare it with a set of ground truth layouts from nuScenes.

To evaluate uncertainty in our diversity results, we use Expected Calibration Error (ECE)~\cite{newer_ECE, ECE_Score}. ECE compares output probabilities to model accuracy. It splits the results into several bins and measures a weighted average discrepancy between accuracy and confidence within each bin: {
\begin{equation}
    \text{ECE} = \sum_b^B n_b | \texttt{acc}(b) - \texttt{conf}(b) |/n,
\end{equation}
where $\texttt{acc}$} is the empirical accuracy, and $\texttt{conf}$ is the estimated confidence for each bin, $n_b$ is the \# of samples per each bin and $n$ is the total \# of samples. ECE is an important metric in BEV map segmentation, as ignoring uncertainties can lead to bad consequences in driving planning. 
We generate 15 diverse outputs for every instance and use the mean as the confidence. The baseline model is trained end-to-end with cross-entropy loss, so the predictions produced by the softmax function are assumed to be the pseudo probabilities.

\myparagraph{Baselines.} 
 For LiDAR-only segmentation, we use PointPillars~\cite{poinpillar}, and BEVFusion-L~\cite{bev_fusion,centerpoint} as our baselines. To the best of our knowledge, BEVFusion-L is the current state-of-the-art method in LiDAR-based BEV map segmentation.

 For camera-only segmentation, our baseline model are LSS~\cite{LSS}, OFT~\cite{OFT}, and BEVFusion-C~\cite{bev_fusion}. Among them, BEVFusion-C is the current state-of-the-art model with a significantly higher IoU score than other methods. We use open-source code from BEVFusion. We also provide results of multi-modal models~\cite{bev_fusion,PointPainting,MVP} for reference.

\begin{table}[!t]
\caption{Ablation studies for MapPrior on nuScenes.}
\setlength{\tabcolsep}{4pt}
\small\centering
\begin{tabular}{cccc}
\toprule    
Output Loss $\mathcal{L}_\mathrm{out}$ & $T(\bz^\prime)$ or $T(\bz^\prime, \bx)$  & mIoU ($\uparrow$) & MMD ($\downarrow$) \\
\midrule
-- & $T(\bz^\prime)$ & 45.6 & \textbf{33.3} \\
\checkmark & $T(\bz^\prime)$ & 49.0 & 45.6 \\
\checkmark & $T(\bz^\prime, \bx)$ & \textbf{51.1} & 35.8 \\
\bottomrule
\label{tab:abalation}
\end{tabular}
\end{table}

\myparagraph{Implementation details.} 
Following \cite{bev_fusion}, we perform segmentation in a [-50m, 50m]$\times$[-50m, 50m] region around the ego car with a resolution of 0.5 meters/pixel, resulting in a final image size of $200\times200$.
Since map classes may overlap, our model performs binary segmentation for all classes. 
The encoder and decoder comprise four downsampling and four upsampling blocks processing a series of 128-256-512-256-128 channels.
Each block comprises $2$ resnet blocks and one convolution and uses sigmoid activation and GroupNorm. The resolution of the latent space is [12,12].

For the generative step, we train a minGPT \cite{mingpt} transformer conditioned on the generated sequence, extracted BEV features, and the initial noisy map. We train the whole model using {Adam~\cite{adam}} with a learning rate of 9.0e-6.
The transformer's BEV feature encoder for the transformer has a similar structure to the model encoder consisting of 3 down-sampling blocks. 
The BEV feature encoder converts the original BEV features shaped [128, 128] into latent space tokens shaped [12, 12]. We apply a multiplier of 100 on the output loss to balance the magnitude of different losses.

\begin{figure*}[!t] \centering  
\includegraphics[width=0.85\textwidth]{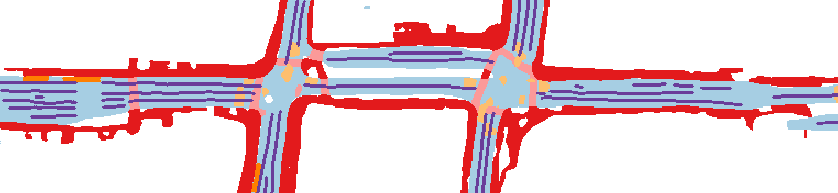} 
\caption{Results for perpetual traffic scene generation. Two road subsections are shown here. } 
\label{Fig4}
\vspace{-3mm}
\end{figure*}

\subsection{Map Segmentation as Generation}
\label{sec:4.2}
\paragraph{Quantitative results.}
We show our quantitative results for map segmentation in Tab.~\ref{tab:IOU}. The results show that MapPrior achieves state-of-the-art performance. Comparing with BEVFusion-L, our MapPrior-L offers \textbf{2.5\%} improvement in mIoU, which is brought purely by our proposed generative stage.
Furthermore, MapPrior provides a substantial improvement in the MMD score compared to baselines. MMD is a metric of distance between the generated layout predictions and the ground truth distribution. This shows that MapPrior's outputs closely match the ground truth data distribution. In addition, this stark difference in MapPrior's MMD performance compared to the baselines implies that the realism metric and precision metric are not closely coupled. 
It is possible to achieve higher IoU while generating non-realistic samples, or vice versa. 
Our approach simultaneously pushes the limit of the two. 

{Moreover, inference speed is vital for MapPrior. Using an RTX A6000 GPU, we compared the inference speeds of our model and BEVFusion in terms of frames per second (FPS). Our findings are presented in Tab.~\ref{tab:one-step}. These results indicate that one-step MapPrior is significantly closer to real-time performance compared to the standard MapPrior, with only a minor trade-off in IoU for increased uncertainty awareness.} 

\begin{table}[!t]
\caption{{Performance of one-step MapPrior.}}
\setlength{\tabcolsep}{4pt}
\small\centering
\begin{tabular}{lcccc}
\toprule    
 & Modality & mIoU ($\uparrow$) & MMD ($\downarrow$) & FPS \\
\midrule
BEVFusion-C~\cite{bev_fusion} & C & 56.6  & 39.6 & 8.85\\
MapPrior-C & C & \textbf{56.7} & \textbf{28.4} & 0.60\\ 
MapPrior-C (1 step)  & C &  {56.7} & {28.7} & 4.26\\ 
\midrule
BEVFusion-L~\cite{centerpoint,bev_fusion}     & L & 48.6 & 115.2 & 7.52\\ 
MapPrior-L & L & \textbf{51.1} & \textbf{35.8}  & 0.57\\
MapPrior-L (1 step) & L  & 50.0 & {50.2} & 4.88\\
\midrule
{BEVFusion~\cite{bev_fusion}} & C+L & {62.7} & \textbf{21.6} & 5.52\\
MapPrior-CL & C+L & \textbf{63.1} & 28.0 & 0.55\\
MapPrior-CL (1 step) & C+L & 63.0 & 28.1  & 3.61\\
\bottomrule
\label{tab:one-step}
\end{tabular}
\vspace{-5mm}
\end{table}

\myparagraph{Qualitative results.}
We show our qualitative results in Fig.~\ref{qual_results}. Compared to other methods, MapPrior can consistently generate coherent and realistic class predictions within the entire map region. Our model has a more coherent divider layout, whereas the baseline methods usually have broken/missing lane dividers. In the baseline methods, the stopline is often jagged and appears disconnected from the road. In distant areas from the ego car, our model tries to make a plausible layout estimate when other methods fail due to limited observations (this is especially noticeable in the pedestrian walkways). The edges from our method are also more smooth, resulting in a more realistic estimation.

\myparagraph{Diversity and uncertainty calibration.}
We show that our model can produce a better-calibrated layout with diversity. In Fig.~\ref{Sampling Diversity}, we show that our model can generate multiple diverse and feasible layouts, All of which are realistic.

By aggregating the diverse samples, our model can produce a calibrated uncertainty map. We show our results in Fig.~\ref{Uncertainty Calibration}. Compared to the baselines (which are unable to generate multiple samples), our uncertainty map aligns with the error map much more accurately.
ECE scores in Tab.~\ref{tab:IOU} further validate this quantitatively.

\myparagraph{Perpetual generation.}
We show our qualitative results for generating 'infinite' roads in Fig.~\ref{Fig4}. We have generated a single 30km long road. Due to size constraints, we are only able to show a subsection. The generated traffic scene contains a highway with intermittent intersections resembling a road layout in a city. We provide the entire road as a gif in our supplementary materials.

\subsection{Discussions}

\paragraph{Ablation studies.}
To justify our design, we provide ablation studies in Tab.~\ref{tab:abalation}. L2 loss at the output end changes the model's optimization target. It puts more weight on generating accurate results as opposed to generating i.i.d. data. Therefore, the model achieves a higher IoU score at the cost of a slightly worse MMD. Providing the transformer with BEV features boosts the performance by around 3\% in the IoU score, and decreases MMD by around 21\%. 

\myparagraph{Inference speed.}

{Inference speed is essential for MapPrior, especially given the real-time demands of autonomous driving. While transformers offer remarkable generative abilities, they inherently slow down performance due to their sequential token generation process. Moreover, the need for MapPrior to produce diverse outcomes further curtails its inference speed. To address this, we introduced the one-step variant of MapPrior, which predicts a single sample for each input and generates all tokens simultaneously. As evidenced in Tab.~\ref{tab:one-step}, the one-step MapPrior registers a marginally worse MMD score and falls short in uncertainty awareness. Nonetheless, the one-step MapPrior markedly outpaces the standard MapPrior in speed.}
\section{Conclusion}

This paper presents MapPrior, a novel generative method for performing BEV perception. The core idea is to leverage a learned generative prior over traffic layouts to provide diverse and accurate layout estimations, which potentially enable more informed decision-making and motion planning.  Our experiments show that our approach produces more realistic scene layouts, enhances accuracy, and better calibrates uncertainty compared to current methods.

\paragraph{Acknowledgement.}
The project is partially funded by the Illinois Smart Transportation Initiative STII-21-07, an Amazon Research Award, an Intel Research Gift, and an IBM IIDAI grant. We also thank NVIDIA for the Academic Hardware Grant. VZ is supported by New Frontiers Fellowship.

{\small
\bibliographystyle{ieee_fullname}
\bibliography{egbib}
}

\end{document}